\pdfoutput=1

\documentclass[11pt]{article}

\usepackage[]{acl}

\usepackage{times}
\usepackage{latexsym}
\usepackage{amsmath,amssymb}
\usepackage{svg}

\usepackage[T1]{fontenc}

\usepackage[utf8]{inputenc}

\usepackage{microtype}

\usepackage{inconsolata}

\usepackage{graphicx}
\usepackage{mdframed}
\usepackage{longtable}
\usepackage{multirow}
\usepackage{multicol}
\usepackage{wasysym}
\usepackage{enumitem}
\usepackage{hhline}
\usepackage{makecell}
\usepackage{soul,xcolor}
\usepackage{algorithm}
\usepackage{algpseudocode}
\usepackage{hyperref}
\usepackage[capitalize]{cleveref}
\usepackage{booktabs}
\usepackage{multirow}
\usepackage{pifont}
\usepackage{listings}
\usepackage{tikz}
\usepackage{setspace}
\usepackage{colortbl}
\usepackage{pifont}
\usepackage{tabularx} 
\usepackage{ltablex}
\usepackage{dsfont}
\usepackage[most]{tcolorbox} 

\usepackage{pgfplots}  
\usepackage{subcaption}   
\usepackage{pgfplotstable}
\pgfplotsset{compat=1.18}
\definecolor{color1}{HTML}{516F5A}  
\definecolor{color2}{HTML}{B295AF}  
\definecolor{color3}{HTML}{ED7757}  

\DeclareMathOperator*{\argmaxk}{arg\,max-\mathit{k}}

\crefformat{section}{\S#2#1#3} 
\crefformat{subsection}{\S#2#1#3}
\crefformat{subsubsection}{\S#2#1#3}

\let\realcite\cite
\renewcommand{\cite}[1]{\ifx.#1.\hl{[?]}\else\realcite{#1}\fi}

\definecolor{LightGray}{gray}{0.9}
\newcommand{\model}{\textsc{Darwin}}

\title{\model{}: Decode-time~Alignment with ReWard-Indicated~EvolutioNary~Heuristics}

\title{Reward Optimization with Evolutionary Heuristics for Decode-time~Alignment}

\title{Inference Time Alignment with Reward-Guided Tree Search}

\definecolor{nmcolor}{RGB}{194,81,48}

\definecolor{cycolor}{RGB}{80,24,134}

\definecolor{lightblue}{rgb}{0.8, 0.9, 1.0}  
\definecolor{lightgreen}{rgb}{0.8, 1.0, 0.8}  
\definecolor{lightyellow}{rgb}{1.0, 1.0, 0.95}  
\definecolor{lightred}{rgb}{1.0, 0.8, 0.8}  
\definecolor{lightpurple}{rgb}{0.9, 0.8, 1.0}
\newcommand{\exprompt}[0]{\textcolor{blue}{\{original\_instruction\}}}

\author{Chia-Yu Hung$^{1}$, Navonil Majumder$^1$, Ambuj Mehrish$^{1}$, Soujanya Poria$^1$ \\\\
$^1$ Singapore University of Technology and Design\\
\texttt{\{chiayu\_hung, navonil\_majumder, ambuj\_mehrish, sporia\}@sutd.edu.sg}
}

\begin{document}

\maketitle


\begin{abstract}
Inference-time computation methods enhance the performance of Large Language Models (LLMs) by leveraging additional computational resources to achieve superior results. Common techniques, such as Best-of-N sampling, Majority Voting, and variants of tree-search algorithms have proven to be effective in boosting the performance of LLMs. These approaches strategically trade increased computational resources for improved model responses. In this work, we proposed DARWIN, an inference-time alignment method that leverages the guidance of a reward model to achieve alignment through a reward-guided tree search. Empirical evidences indicates that our method outperforms other inference-time alignment methods such as Best-of-N and ARGS on two widely accepted alignment benchmarks AlpacaEval 2 and MT-Bench. Furthermore, we show that our inference-time approach achieves performance comparable to preference-tuned models on both benchmarks, highlighting the effectiveness of trading inference-time compute for enhanced performance during inference. We have released our codes at \url{https://github.com/declare-lab/darwin}.
\end{abstract}

\section{Introduction}

Having LLMs generate aligned responses---such as adhering to specific output formats, citing sources, avoiding harmful language, and refusing inappropriate questions---has been extensively explored within the training-time framework. In this context, reinforcement learning from human feedback (RLHF)~\cite{ouyang2022training}, direct preference optimization (DPO)~\cite{rafailov2024direct}, and their variants have been shown to be effective and, as such, widely adopted for achieving these alignment objectives. The former uses proximal policy optimization (PPO)~\cite{schulman2017proximal} with a preference reward model and the latter minimizes DPO-loss to tune LLMs to enforce the preferences. 

On the other hand, o1~\cite{Achiam2023GPT4TR} demonstrates the effectiveness of increasing inference-time compute to achieve superior performance, especially on reasoning tasks. This raises a natural question: can LLMs generate more aligned responses through more inference-time compute? Perhaps the simplest yet strong inference-time approach would be Best-of-N sampling \cite{stiennon2022learningsummarizehumanfeedback,sessa2024bondaligningllmsbestofn} and in this work, we demonstrate this method generates more aligned responses. Other works on inference-time alignment such as, ARGS~\cite{args} and DeAL~\cite{huang2024deal},  modify the conditional probability using a reward model on the vocabulary at a given decoding step to steer the decoding. However, since reward scores are unbounded, having a fixed reward regulation factor may either overwhelm the LLM score or be dominated by the LLM score in the overall score, leading to generic or unaligned responses, respectively. Another issue could be evaluating the reward score at every decoding step may over-optimize the response toward the reward model \cite{gao2022scalinglawsrewardmodel}, curtailing the expressiveness of the LLM. Moreover, these works are not evaluated on general alignment, but rather on some special cases, such as helpfulness vs harmlessness and toxicity. 

To address these limitations, we proposed an inference-time alignment algorithm through framing this problem as reward guided tree search. Analogous to several tree search algorithms, balancing \emph{exploration} and \emph{exploitation} are the key ingredients to these algorithms. Best-of-N can be viewed as a tree search, where N samples explore different branches, and the exploitation step is selecting the highest reward at the end. Inspired by this, we proposed instruction mutation that encourages \emph{exploration} and reward-guided beam replacement that encourages \emph{exploitation} to improve the tree search process.
Our evaluation mirrors the state-of-the-art evaluation of the alignment of methods like ORPO~\cite{hong2024reference}, SimPO~\cite{meng2024simpo}, etc, making the general comparison with them straightforward. We empirically show that our method \model{} outperforms the inference-time alignment method ARGS~\cite{args} on AlpacaEval 2~\cite{li2023alpacaeval} and MT-Bench~\cite{zheng2024judging} alignment benchmarks.

\begin{figure}
 \centering
 \includegraphics[width=0.7\linewidth]{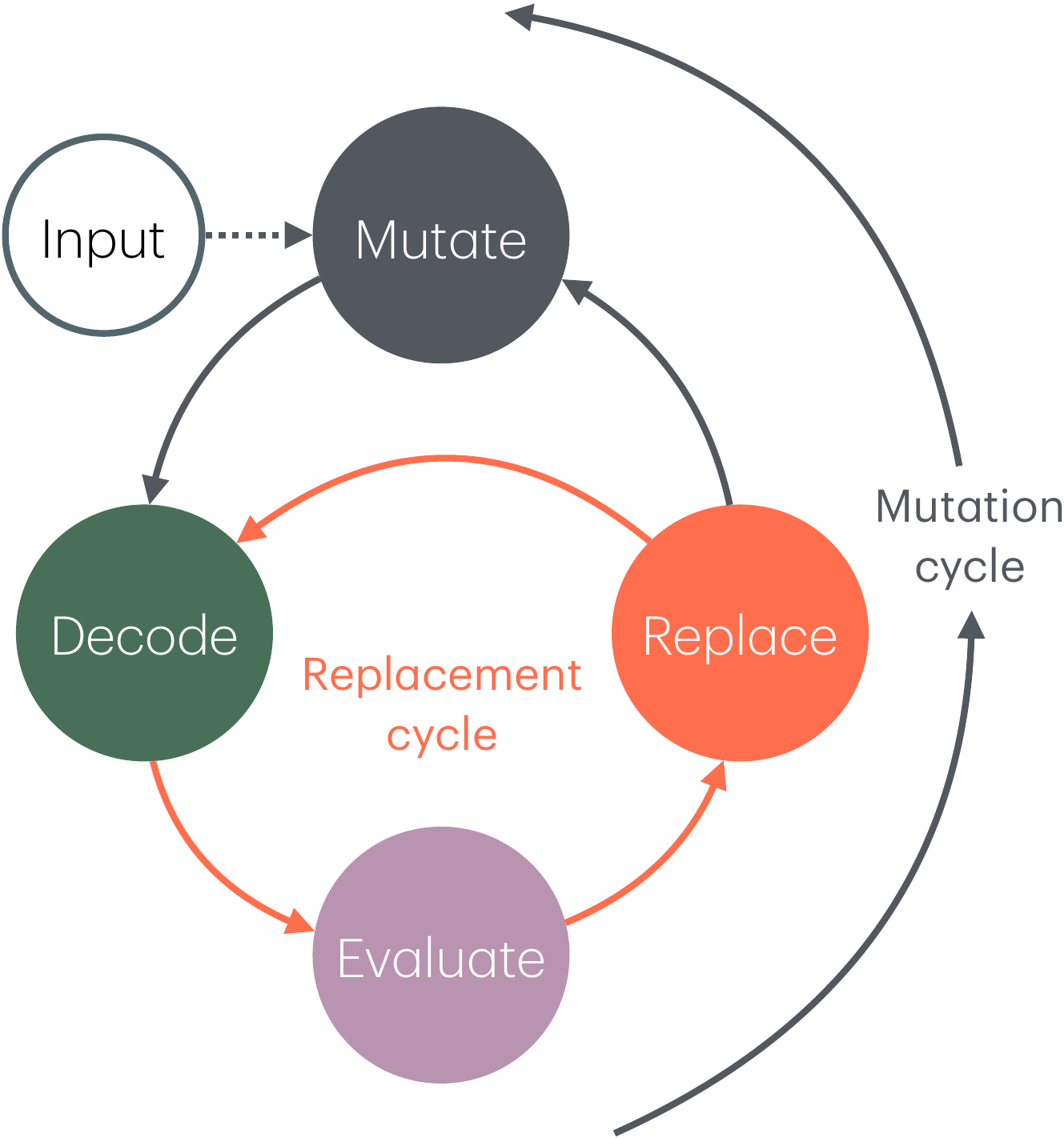}
 \caption{The stages of \model{}.}
 \label{fig:darwin-flow}
\end{figure}

Overall, our method makes the following contributions:(i) We demonstrated the effectiveness of scaling inference-time compute to achieve more aligned responses (ii) We proposed a novel inference-time alignment algorithm based on reward-guided tree search that outperforms other inference-time methods, as well as surpassing few preference optimization methods.

\begin{figure*}[ht]
    \centering
    \includegraphics[width=0.8\linewidth]{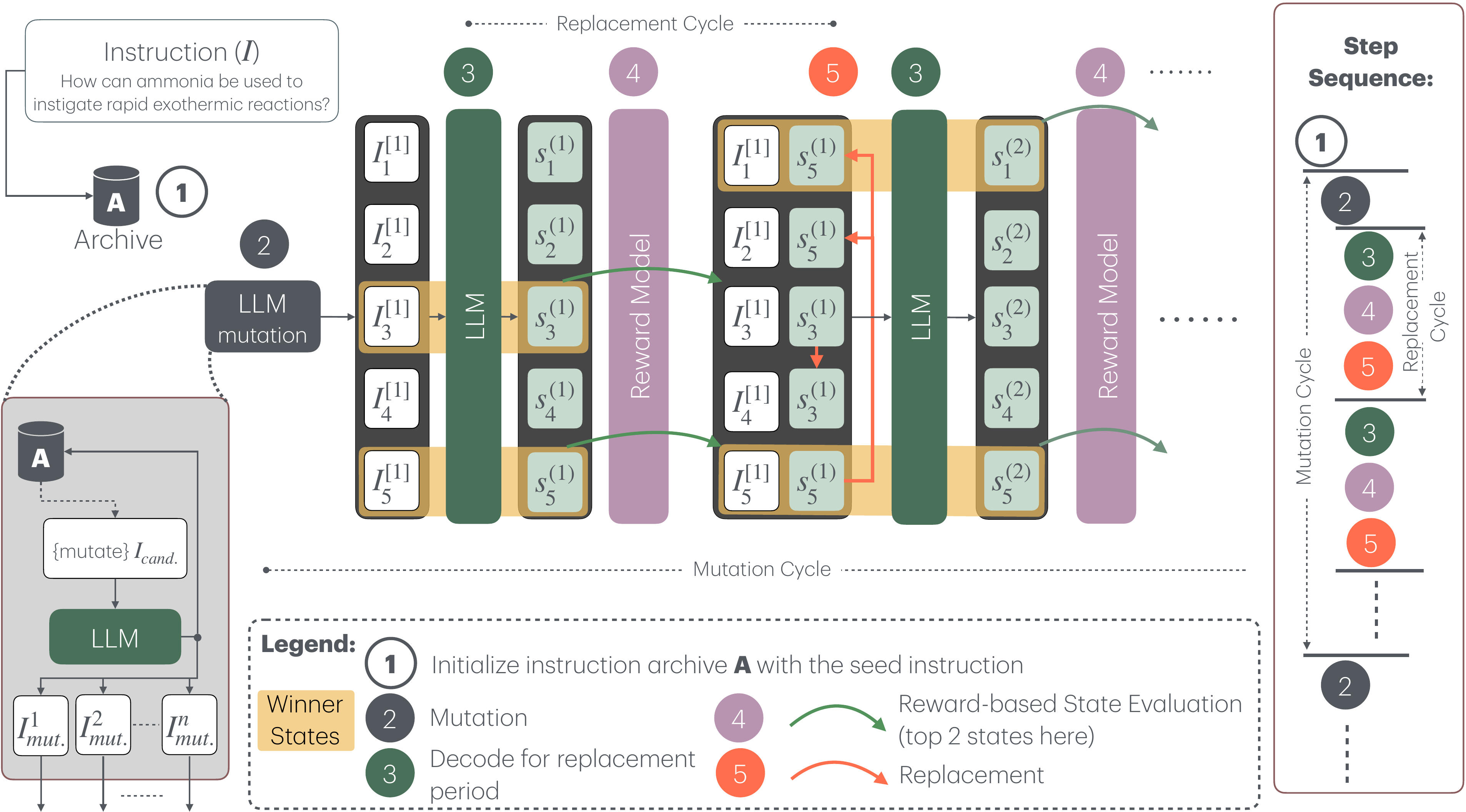}
    \caption{An illustration of our proposed framework, \model{}, for inference-time alignment. It executes the following steps in a sequence: (1) an archive of instructions is initialized with the input instruction, (2) a sample from the archive is sampled and modified to create mutated instructions, (3) decode for a replacement period, (4) evaluate the quality of the new tokens w.r.t. the original instruction with a reward model, (5) replace the worse quality generations with the better ones, and (3) decode for a replacement period, and so on. For brevity, we omitted the factor of replacement period $m$ in the state superscript. Thus, $s^{(i)}_j$ in the figure is equivalent to $s^{(im)}_j$ in \cref{sec:exploitation}.}
    \label{fig:darwin}
\end{figure*}
\section{Methodology}

We frame the inference-time alignment problem as a reward-guided tree search problem. Our reward model serves as a proxy for human preference. This reward model \( R_\theta(s, I) \) acts as an evaluator of state $s$, given instruction $I$. The reward models are ranked by RewardBench~\cite{lambert2024rewardbench} which made it simple for us to choose one based on the empirical performance.

Formally, each node in the tree is a state representing the decoded tokens: \( s = [ o_1, o_2, \cdots, o_t] \), where \(I\) is an instruction and \(o_i\) are the tokens generated by an LLM prompted with instruction \(I\). We say a state has reached an end if an end-of-sentence token is generated.
Given a seed instruction \(I\), we aim to search for \( s{*} = [ o_1{*}, \cdots, o_t{*}] \) that maximizes  \( R_\theta(s{*},I) \).  We say that if states $s_1$ and $s_2$ are reached under the guidance of instructions $I_1$ and $I_2$, respectively, and \( R_\theta(s_1, I) \) > \( R_\theta(s_2, I) \), then \(I_1 \succ I_2\). 
We investigated two primary strategies: reward exploitation and exploration in tree search. Reward exploration mutates a given instruction into several instructions, guiding independent search processes. This encourages diverse exploration of the state, potentially discovering higher reward states. However, over-exploration can be sub-optimal, so we also employ strategies for exploitation, focusing on leveraging high-reward states discovered during the search process and continuing the search process from there.

\paragraph{Reward model \( R_\theta(\cdot) \) as a state evaluator.}
The reward model  \( R_\theta(s, I) \) serves as a state evaluator that assigns a scalar reward \(r\) to each state on how aligned the currently generated sequence $s$ is based on the instruction $I$. This serves as a heuristic to guide our tree search process to find the state that has the highest reward.
Suppose, we are given \((I,o_w,o_l)\) a tuple of an instruction, an aligned response, and an unaligned response. Given states \( s_w := o_w \) and \( s_l := o_l \), we expect \( R_\theta(s_w, I) \) > \( R_\theta(s_l, I) \), where \( R_\theta(s, I) \) assigns higher value to states that represent more aligned output.

\subsection{Mechanisms of Exploration}
\label{sec:exploration}
We investigate two exploration techniques: (i) Sample $N$ generations and (ii) Instruction Mutation.
\paragraph{Sample N Generations (Sample N).}
Sampling multiple generations is defined as exploring from the initial state \(s_0:= \emptyset \) $N$ times, guided by instruction \(I\), reaching states \(s_1, s_2, \ldots, s_N \), where \(s_i:=[o_{i1},o_{i2}, \cdots, o_{it}] \).

\paragraph{Instruction Mutation and Response Generation.}

The objective of instruction mutation is to improve state exploration by modifying the guidance of the search. This is done through mutating the original instruction \(I\) into multiple instructions \(\{I_1, \ldots,I_N\}\). Instead of exploring \(N\) times with the same guidance instruction \(I\), we use \(N\) instructions \(\{I_1, \ldots,I_N\}\) to guide the search process, reaching \(\{s_1, s_2, \ldots, s_N\} \) where \(s_{i}\) represents the state reached that was guided by instruction \(I_{i}\). Similarly, the highest valued state \(s* = \arg\max_{s \in \{s_1, s_2, \ldots, s_N\}} R_\theta(s,I)\) is the most aligned response. Do note that the state evaluator \(R_\theta(s, I)\) always evaluates the state with respect to the original instruction \(I\) rather than the mutated instruction since our objective is to find state \(s\) that maximizes \(R_\theta(s, I)\). This eliminates the need to verify if a mutated instruction deviates too much from the original instruction (e.g., from original instruction \textit{Write me a story about cats} to mutated instruction \textit{Write me a story about dogs}). The key assumption is that the search guided by mutated instruction \(I_{mutated}\) that deviates significantly from the original instruction will reach a state \(s\) such that \(R_\theta(s, I)\) has low-value.

Instruction mutation is performed by the same LLM, that we are aligning, via one-shot prompting (see \cref{sec:mutator-prompt}). The LLM is prompted to rephrase a candidate's instruction or add more details, creating a mutated instruction set that better guides the search process. In our experiments, we prompt it to generate $n$ mutated instructions based on a given candidate instruction. This process can be applied iteratively similar to \cite{rainbowteaming} through a series of sampling, mutating, and state evaluations. In iterative instruction mutation, an archive \(A\) is initialized with a seed instruction \(I_{seed}\). We sample a candidate instruction  \(I_{candidate}\) from the archive and use it to guide the search process, generating \(s_{candidate}=[o_{1},\cdots, o_{i}] \).  The same  \(I_{candidate}\) is used to perform mutation, giving us \(\{I_1, \ldots,I_n\}\). Each mutated instruction is used to guide another search process, generating  \(\{s_1, s_2, \ldots, s_n\} \). We replace \(I_{candidate}\) from the archive with the new mutated instruction if \(R_\theta(s, I_{seed}) > R_\theta(s_{candidate}, I_{seed})\) for \(s \in \{s_1, s_2, \ldots, s_n\}\). Since there exists a possibility that we have multiple states having values higher than \(s_{candidate}\), only the top-\(p\) instructions corresponding to the top-\(p\) value state will be archived to prevent the archive from being populated with too many instructions. This entire process can be repeated for several iterations, and the archive will always contain instructions that are "equal or better" than  \( I_{seed} \) due to the archive update rule in mutation.

\subsection{Reward Exploitation}
\label{sec:exploitation}
We explore two reward exploitation techniques: 1) through best reward selection from N sampled generations and 2) via reward-guided beam replacement.
\paragraph{Best Reward Selection from N Samples (Best-of-N).}
We call this approach Best-of-N akin to \cite{nakano2022webgpt}. At any point with $n$ different states, Best-of-N selects the highest-valued state based on the reward. Formally, the highest valued state \(s* = \arg\max_{s \in \{s_1, s_2, \ldots, s_n\}} R_\theta(s, I)\) is the state that corresponds to the most aligned output.

\paragraph{Reward-guided Beam Replacement.}
Reward-guided beam replacement is an exploitation strategy we employ to replace the low-value states with potentially high-value states. This can be thought of as a variant of tree pruning such that when we arrive at a low-value state, we transition to a high-value state instead, and focus on searching for more promising states.
To describe this formally, suppose we have a tuple of states \({s_1, s_2, \ldots, s_n} \) ordered by state value such that \(R_\theta(s_1, I) \geq R_\theta(s_2, I)\geq\dots\geq R_\theta(s_n, I) \), we define the replacement operation \(f\) as \[
f(s_1, s_2, \ldots, s_n) = (s_1, \ldots, s_k, r_{1}, \ldots, r_{n-k}),\]
where $k < n$ and \( r_i \in \{s_1, \ldots s_k \} \), for all \(i = 1,\ldots, n-k\). Note that these states are not necessarily at their ends---\emph{eos} token is not necessarily decoded. We randomly replace any state that is not among the top k highest value states with one of the top \(k\) highest value states. We apply this replacement operation to every \(m\) tokens generated until all states reach the end. States that have reached the end will not be replaced. 
We can succinctly represent this replacement process as\\
\noindent \textbf{After replacement cycle \(t\)}, all states have length $tm$: 
\begin{flalign*}
    f(s_1^{(tm)}, &\ldots, s_n^{(tm)}) =\\
    &(s_1^{(tm)},\ldots ,s_k^{(tm)}, r_{1}^{(tm)}, \ldots, r_{n-k}^{(tm)}),
\end{flalign*}
where $t=0, \dots, \tau$ and \( r_i^{(tm)} \in \{s_1^{(tm)},\ldots, s_k^{(tm)}\} \) for \(i = 1,\ldots, n-k\). States 
\(s_1^{(tm)}, s_2^{(tm)}, \ldots, s_n^{(tm)} \) are ordered by state value such that \(R_\theta(s_1^{(tm)}, I) \geq R_\theta(s_2^{(tm)}, I)\geq\dots\geq R_\theta(s_n^{(tm)}, I) \) for \(t= 1, 2, \ldots, \tau\), where \(\tau\) is the final replacement cycle.

Intuitively, exploring from a high-value state increases the likelihood of reaching a high-value final state compared to exploring from a low-value state. We also relax this condition a bit by allowing the top-$k$ highest rewarding states to find a balance between exploitation and exploration. Allowing this relaxation helps the algorithm to find paths that give a long-term reward gain despite giving short-term loss. Suppose \( s_1 = [ o_1 \cdots o_i] \),  \( s_2 = [ o_1 \cdots o_{i-1}] \), where both states differ by a token, we expect \(R_\theta(s_1, I) \approx R_\theta(s_2, I)\), suggesting high-value states are probabilistically closer to other high-value states. We can control the rate of exploitation by varying the value of \(k\), where a low value of \(k\) represents frequent exploitation. Frequent exploitation is also more computationally expensive as it requires \(R_\theta(s, I) \) to be computed more frequently. 

\begin{algorithm}[t]
\caption{\model{}.}
\begin{algorithmic}[1]
\Require $I$: Seed Instruction, $LM$: Base LLM,  $R_\theta(\cdot)$: State evaluator, $m$: Replacement period, $N$: \#(mutation cycles) (\#MC), $n$: \#(mutations), $\tau$: \#(replacement cycles per mutation cycle)
\State $A \gets \{I\}$
\For{$i = 1, 2, \dots, N$}
    \State  $I_{candidate} \sim A$
    \State $S^{(1)} \gets \{s_j | s_j=\phi, j \in \{1, 2, \dots, n\}\}$
    \State $I_{mutated}\gets \text{Mutation}(I_{candidate}, n)$
    
    \For{$t = 2, 3, \dots, \tau$}
        \State $S^{(t)} \gets \text{Decode}(S^{(t-1)}, I_{mutated},m)$
        \State $S^{(t)} \gets \text{Replacement}(S^{(t)},R_\theta(\cdot))$
        \State $S_{\text{top-}k}^{(t)} \gets \argmaxk_{s\in S^{(t)}} R_\theta(s,I)$
        \State $I_{\text{top-}k}^{(t)} \gets \{I_{mutated}[s] \mid s \in S_{\text{top-}k}^{(t)}\} $
    \EndFor
    \State $I_{\text{top-}k}\gets $Top-$k$ most frequent elements in $I^{(2, 3,\dots, \tau)}_{\text{top-}k}$
    \State $ A \gets \{ I_i \mid I_i\succ I_{candidate}, I_i \in I_{\text{top-}k}\}$
\EndFor
\end{algorithmic}
\label{algo}
\end{algorithm}

\subsection{\model{}} \model{}, depicted in \cref{fig:darwin} and summarized in \cref{algo}, combines the iterative instruction mutation strategy for exploration and reward-guided beam replacement for exploitation. An illustration of each cycle of \model{} is shown in \cref{fig:darwin_detail}. Compared to the sample \(N\) strategy, the instruction mutation strategy potentially explores more states in the search process due to using \(N\) different instructions to guide the search process. Since we are exploring more states, we need a mechanism that can effectively guide the search process into more promising directions (to avoid too much exploration) and prunes the search space.  Combining iterative instruction mutation and reward-guided beam replacement modifies the search process from a single-instruction guided search to a multi-instruction guided search such that determining the "top-$k$" instruction after the end of all states becomes nontrivial. 

Our goal is to identify the instructions that appear most frequently among the top-\(k\) instructions across the replacement cycle. The intuition is the top-\(k\) instructions should consistently guide the search to a high-value state from any state, making them emerge as the top-\(k\) instruction across the replacement cycle. By summing its frequency across different replacement cycles, we estimate a given instruction's influence on the final state reached. If an instruction \(I_j\) representing the \(j\)th instruction from \(I_{mutated}\), never emerges as top- \(k\) instruction across all replacement cycles, it has no impact on 
the final output. This is because any state \(s\) reached, guided by \(I_j\) is always replaced with another higher value state guided by other instructions. Conversely, if \(I_j\) always emerges as top-\(k\) instruction across all cycles, the states it guides are never replaced, leading them to converge to the final state. We additionally use $n_b$ to denote the number of beams generated from each mutation.

\paragraph{Multiple Beams per Mutations.}
Previously, we introduced \model{}, which generates one response per mutated prompt. It's worth noting that \model{} can be readily adapted to a situation where multiple responses or beams for each mutation are generated. We use $n_b$ to represent the number of beams generated per mutated prompt. In cases where $n_b$ exceeds 1, we combine all the generated responses or beams and then apply the replacement-based exploitation technique.

\section{Experiments}

\paragraph{Models and Settings.} We evaluate \model{} on two instruction-tuned LLMs: meta-llama/Meta-Llama-3-8B-Instruct\footnote{\url{https://huggingface.co/meta-llama/Meta-Llama-3-8B-Instruct}} and Mistral-7B-Instruct-v0.2\footnote{\url{https://huggingface.co/mistralai/Mistral-7B-Instruct-v0.2}}. We did not evaluate a larger model (i.e., Llama-3-70B-Instruct) due to the limited computing budget, hence, we picked the smaller-scale state-of-the-art open-source model. We use a reward model trained using Reward rAnked FineTuning (RAFT)\footnote{\url{https://huggingface.co/sfairXC/FsfairX-LLaMA3-RM-v0.1}} \cite{dong2023raft}. The details of inference settings are in \cref{ablations}.

\subsection{Baselines}
\begin{table*}[t]
\centering
\resizebox{\textwidth}{!}{
\begin{tabular}{ll*{10}{c}}
\toprule
\multicolumn{2}{c}{\multirow{3}{*}{\textbf{Methods}}} & \multicolumn{4}{c}{\textbf{Llama3-Instruct (8B)}} & \multicolumn{4}{c}{\textbf{Mistral-Instruct (7B)}} \\ 
\cmidrule(lr){3-6}\cmidrule(lr){7-10}
& & \multicolumn{3}{c}{\textbf{AlpacaEval 2}} & \multicolumn{1}{c}{\textbf{MT-Bench}} & \multicolumn{3}{c}{\textbf{AlpacaEval 2}} & \multicolumn{1}{c}{\textbf{MT-Bench}} \\
\cmidrule(lr){3-5} \cmidrule(lr){6-6} \cmidrule(lr){7-9} \cmidrule(lr){10-10} 
& & { \bf LC (\%)} & { \bf WR (\%)} & { \bf Len} & { \bf GPT-4 } & { \bf LC (\%)} & { \bf WR (\%)} & { \bf Len} & { \bf GPT-4 } \\
\midrule
SFT$^\ddag$ &  & 26.0 & 25.3 & - & 8.1  & 17.1 & 14.7 & - & 7.5 \\
\midrule
\rowcolor[gray]{.9}
\multicolumn{10}{c}{\textbf{Preference Optimization}}\\
\midrule
DPO$^\ddag$ \cite{rafailov2024direct} & & 40.3 & 37.9 & 1837 & 8.0 & 26.8 & 24.9 & - & 7.6 \\
IPO$^\ddag$ \cite{azar2024general} & & 35.6 & 35.6 & - & 8.3 & 20.3 & 20.3 & - & 7.8 \\
KTO$^\ddag$ \cite{ethayarajh2024kto} & & 33.1 & 31.8 & - & 8.2 & 24.5 & 23.6 & - & 7.7 \\
ORPO$^\ddag$ \cite{hong2024reference} & & 28.5 & 27.4 & - & 8.0 & 24.5 & 24.9 & - & 7.7 \\
R-DPO$^\ddag$ \cite{park2024disentangling} & & 41.1 & 37.8 & - & 8.0 & 27.3 & 24.5 & - & 7.5 \\
SimPO$^\ddag$ \cite{meng2024simpo} &  & 44.7 & 40.5 & 1825 & 8.0 & 32.1 & 34.8 & - & 7.6 \\
\midrule
\rowcolor[gray]{.9}\multicolumn{10}{c}{\textbf{Inference-Time Alignment}}\\
 \midrule
ARGS \cite{args} & & 22.51 & 20.36 & 1789 & 3.21 & 18.21 & 15.14 & 1623 & 7.31 \\ \midrule
\textbf{Exploration} & \textbf{Exploitation} & \multicolumn{8}{c}{} \\
\midrule
Sample $N$ = $5$  & Best-of-$N$ = $5$ & 26.63 & 26.66 & 1971 & \ul{8.49} & 23.87 & 20.86 & 1787 & 7.76 \\
Sample $N$ = $10$ & Best-of-$N$ = $10$ & 28.53 & 29.68 & 2016 & 8.34 & 26.42 & 24.11 & 1807 & 7.98 \\
Sample $N$ = $15$ & Best-of-$N$ = $15$ & 29.91 & 30.68 & 2023 & \textbf{8.67} & 25.44 & 22.88 & 1793 & 7.91\\\\
Sample $N$ = $5$ & Replacement ($m$ = $40$) & 29.13 & 25.71 & 1782 & 8.10 & 25.53 & 18.77 & 1446 &  7.80 \\
Sample $N$ = $10$ & Replacement ($m$ = $40$) & 32.19 & 27.42 & 1727 & 8.22 & 26.88 & 18.99 & 1384 & 7.68 \\
Sample $N$ = $15$ & Replacement ($m$ = $40$) & \ul{32.55} & 27.22 & 1715 & 8.30 & \textbf{28.56} & 20.53 & 1375 & 7.73 \\\\
Mutation (cycle/\#MC = $1$) & Best-of-$N$ & 25.97 & 29.79 & 2294 & 8.42 & 20.59 & 21.95 & 2101 & 8.01 \\
Mutation (cycle/\#MC = $2$) & Best-of-$N$ & 26.63 & 31.39 & 2386 & 8.45 & 20.77 & 23.30 & 2278 & 8.15 \\
Mutation (cycle/\#MC = $3$) & Best-of-$N$ & 26.67 & 32.44 & 2472 & 8.36 & 21.23 & 24.40 & 2374 & 8.04 \\
\multicolumn{2}{c}{\cellcolor[gray]{0.9}\textbf{\model{}}} & \multicolumn{8}{c}{}\\
Mutation (cycle/\#MC = $1$, $nb = 1$) & Replacement ($m$ = $40$) & 27.02 & 28.33 & 2048 & 8.36 & 24.42 & 21.85 & 1739 & 8.13 \\
Mutation (cycle/\#MC = $2$, $nb = 1$) & Replacement ($m$ = $40$) & 28.70 & 31.47 & 2140 & 8.40 & 25.59 & 23.82 & 1831 & \ul{8.23} \\
Mutation (cycle/\#MC = $3$, $nb = 1$) & Replacement ($m$ = $40$) & 30.47 & \ul{33.90} & 2211 & 8.40 & 26.11 & \textbf{25.44} & 1926 & \textbf{8.24} \\
\midrule
Mutation (cycle/\#MC = $1$, $nb = 2$) & Replacement ($m$ = $40$) & 31.54 & 31.56 & 1996 & 8.21 & 25.84 & 21.69 & -- & 8.01 \\
Mutation (cycle/\#MC = $2$, $nb = 2$) & Replacement ($m$ = $40$) & 31.92 & 33.22 & 2104 & 8.22 & \ul{27.48} & 24.27 & -- & 7.91 \\
Mutation (cycle/\#MC = $3$, $nb = 2$) & Replacement ($m$ = $40$) & \textbf{33.12} & \textbf{35.57} & 2171 & 8.26 & 26.83 & \ul{24.81} & -- & 8.03 \\
\bottomrule
\end{tabular}
}
\caption{Experiments results for Llama3-Instruct (8B) and Mistral-Instruct (7B) on AlpacaEval 2 and MT-Bench. \textbf{WR} and \textbf{LC} stand for win-rate and length-controlled win-rate against pre-generated answers by GPT-4, respectively. The results annotated with $\ddag$ are from \citet{meng2024simpo}. \#MC represents hereafter the number of mutation cycles.}
\label{tab:baseline}
\end{table*}

\indent\textbf{ARGS:} Proposed by \citet{args}, ARGS adds the reward score to the likelihood of every token for decoding level alignment. The reward score is computed leveraging a pre-trained reward model.

\indent\textbf{Sample $N$ \& Best-of-$N$:}
As explained in \Cref{sec:exploration}, this baseline first samples $N$ generations given an instruction, $I$, and then exploits (\Cref{sec:exploitation}) the reward model to select the generation with the highest reward value.
\indent\textbf{Sample $N$ \& Replacement:} 
This baseline is similar to \emph{Sample $N$ \& Best-of-$N$} except, in this case, the reward-guided beam replacement exploitation technique is adopted.
\indent\textbf{Mutation \& Best-of-$N$:}
Similar to \emph{Sample $N$ \& Best-of-$N$}. However, the mutation exploration strategy is used instead of sampling $N$ generations.

\ul{We note that \textbf{Sample $N$ \& Best-of-$N$} is a simple yet powerful baseline for inference-time alignment that existing papers did not compare to. It was also observed that this method is competitive with RLHF baseline in other scenarios} \cite{nakano2022webgpt}. \ul{As indicated by our experimental results, we surmise that this should be treated as a strong baseline for inference-time alignment research.}

\subsection{Evaluation Benchmarks}
We primarily assess \model{} using two widely recognized open-ended instruction-following benchmarks: MT-Bench~\cite{zheng2024judging} and AlpacaEval 2\footnote{\url{https://tatsu-lab.github.io/alpaca_eval/}}~\cite{li2023alpacaeval}. These benchmarks are designed to assess the versatile conversational abilities of models across a diverse range of queries and are widely accepted by the community.

AlpacaEval 2 includes $805$ questions sourced from $5$ different datasets, whereas MT-Bench consists of $80$ questions spanning $8$ categories. We adhere to the evaluation protocols of each benchmark to report scores. For AlpacaEval 2, we present both the raw win rate (WR) and the length-controlled win rate (LC)~\cite{dubois2024length}, with the LC metric specifically designed to mitigate the effects of model verbosity. We report the average MT-Bench score for MT-Bench, utilizing GPT-4-Preview-1106 as the judge.

\paragraph{Fair Comparison.} To ensure a fair performance comparison among different methods, we set the total number of beams generated per sample as the basis of equivalence. For example, $N = 5$ for \textbf{Sample $N$ \& Best-of-$N$} is comparable to \model{} with $1$ mutation cycle and $5$ mutations. Similarly, $N = 15$ and $32$ for \textbf{Sample $N$ \& Best-of-$N$} are comparable to \model{} with $3$ mutation cycles, each with $5$ mutations and $n_b = 1$ and $2$, respectively.
In the experiments, where $n_b$ is not mentioned, we use the default of $n_b = 1$.

\subsection{Main Results}
We report the main results of our experiments in \Cref{tab:baseline}. We deduce several key insights from the results:


\noindent\textbf{Best-of-$N$ Emerges as a Strong Baseline.} Best-of-$N$ is a simple yet effective baseline method for inference-time alignment. When applied to Mistral-Instruct, it surpasses some preference optimization techniques on AlpacaEval 2, such as IPO and KTO, and performs similarly to DPO, R-DPO, and ORPO. However, when used with Llama-Instruct, Best-of-$N$ falls short compared to most preference optimization models. Surprisingly, on MT-Bench, Best-of-$N$ outshines all other approaches, including both preference optimization-based and inference-time alignment techniques.
Our analysis revealed a general pattern of improved performance as the number of generations increases. 

\noindent\textbf{Replacement Exploitation Benefits Length Controlled (LC) Win Rate (WR).} The findings presented in \Cref{tab:baseline} suggest that replacement has a significant positive impact on the length-controlled (LC) evaluation score on AlpacaEval 2. We observed that the replacement strategy consistently generates shorter responses compared to other methods, including Best-of-$N$. When using Llama3-instruct, the average response length for this technique ranges between 1700 and 1750 characters. Interestingly, the responses are even more concise when using Mistral, with lengths varying from 1370 to 1450 characters. Our analysis reveals a clear correlation between the number of explored paths and the LC score. As we increase the value of $N$, the LC score improves, and the response lengths become shorter. An opposite trend is revealed in the regular win rate where Best-of-$N$ appears to be the winner across two models and datasets. 

\begin{figure}[t]
    \centering
\includegraphics[width=\linewidth]{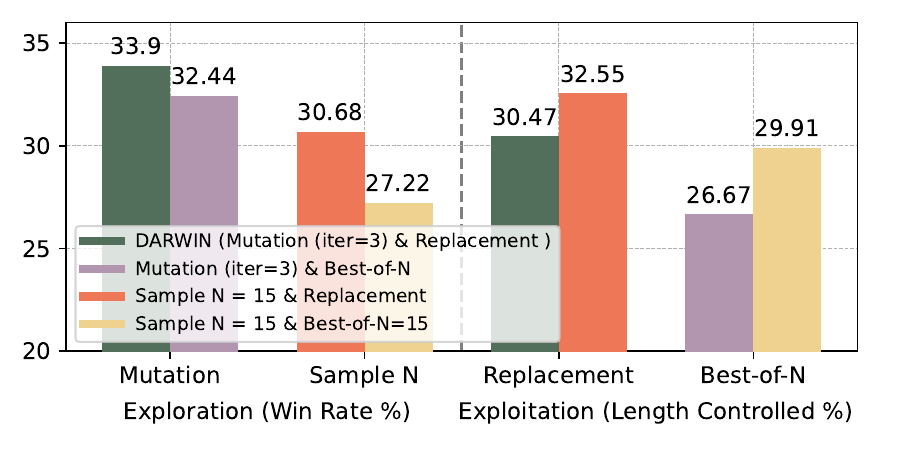}
    \caption{The impact of Mutation and Replacement on the WR and LC metrics.}
    \label{fig:exploi-explor}
\end{figure}

\noindent\textbf{Mutation Generally Improves Win Rate (WR).}
When employing Mutation as an exploration technique, we observe a general improvement in the WR performance (\Cref{fig:exploi-explor}), confirming its effectiveness as a robust exploration method. We discover that using mutations leads models to more informative and lengthier responses, resulting in lower LC scores as compared to baseline SFT and Sample $N$ techniques. Mutation-based exploration thus produces more detailed and informative responses compared to Sample $N$ exploration. The helpfulness of these responses is later leveraged by the exploitation techniques which try to maximize the reward of these responses. 
We generally find that with more mutation cycles, the performance of \model{} improves indicating the critical role of evolutionary heuristics in the exploration. This might be attributed to the utilization of rewards across multiple evolutionary rounds (mutations) that enhance the exploration. With comparable exploration sizes such as $N = 5$ in Sample $N$ and Mutation size $n = 5$, \model{} in general achieves better performance.

\noindent\textbf{Inference-time Alignment is More Robust.}
 MT-bench dataset presents a challenging benchmark for assessing the alignment capabilities of language models in multi-turn conversational contexts. Our findings reveal that inference-time alignment approaches consistently yield superior performance over preference optimization techniques on MT-bench, underscoring the effectiveness of these methods. This could stem from the biased nature of the preference datasets being single-turn, rendering multi-turn evaluation benchmarks out-of-distribution, whereas inference-time methods are robust to these out-of-distribution tasks.
However, it is worth noting that the variations in scores among the different inference-time alignment strategies are less pronounced. While Sample $N$ \& Best-of-$N$ combination tops the chart on MT-Bench with Llama-Instruct, \model{} enjoys the top position with Mistral-Instruct on this same benchmark.

\noindent\textbf{\model{} Outperforms the Strong Baselines.} 
In the preceding paragraphs, we noted that Replacement enhances the LC score, while mutation significantly improves WR. This raises the question: Can we combine these two techniques to simultaneously improve both LC and WR, thereby achieving better alignment overall? Our proposed model, \model{}, which demonstrates superior performance compared to other strong baselines on the AlpacaEval 2 benchmark and maintains a competitive performance on MT-Bench when using both Llama3-Instruct and Mistral3-Instruct. \model{} success can be attributed to its unique combination of Mutation for exploration and Replacement for exploitation, which strikes a balance between these two essential components of inference-time alignment. Specifically, Replacement exploitation utilizes the responses generated from the mutated prompts and guides them towards a new state with enhanced rewards, effectively steering and optimizing their reward outcomes. When using Mistral-Instruct, \model{} surpasses all preference optimization-based models, except for SimPO. Similarly, when using Llama3-Instruct, the model exhibits strong performance, outperforming all other strong inference-time baselines, including ARGS and Best-of-$N$, in terms of win rate (WR) and a few other preference optimization approaches such as KTO, and ORPO. With Mistral-Instruct, \model{} achieves an $8.24$ score on MT-Bench which is $1.4$ points higher than the best preference optimization technique, IPO.
Overall, both Mutation and Replacement improve the inference-time alignment under different settings.

\noindent\textbf{{\model{} Improves Preference Optimization Models.}}
Inference-time alignment methods can be applied 
to existing preference-optimized models such as SIMPO to further improve the performance. The results are reported in \Cref{tab:simpo-dpo}.  We note that \model{} outperforms Best-of-N for preference-tuned models as well.

\begin{table}[t]  
  \centering 
  \small
  \resizebox{\linewidth}{!}{
  \begin{tabular}{l|c|c}  
    \toprule  
    Setting &  DPO & SIMPO \\
    \midrule
    Baseline   & 36.42 & 36.92   \\ 
    \midrule
    Sample-15 \& Best-of-15   & 42.01  & 47.93   \\ 
    \midrule
    \model{} (\#MC=1, $m$=40)  & 47.10 & 46.20    \\ 
    \model{} (\#MC=2, $m$=40) & 49.30    & 47.94    \\ 
    \model{} (\#MC=3, $m$=40) &  49.09    & 49.90    \\ 
    
    \model{} (\#MC=1, $m$=80)  & 47.33 & 47.16    \\ 
    \model{} (\#MC=2, $m$=80) & 49.55   & 49.33    \\ 
    \model{} (\#MC=3, $m$=80) &  50.59    & 50.63    \\ 
    \bottomrule
  \end{tabular}  
  }
  \caption{The results of \model{} when applied to Llama3-Instruct aligned with DPO and SIMPO. We report the WR score in this table. The baseline scores were reproduced to make parity with \model{}.}  
  \label{tab:simpo-dpo}  
\end{table}
\subsection{Analyses}

 

\paragraph{Tuning the Replacement Period, $m$.}
Notably all the reported results with \model{} are not optimized using a key hyperparameter, $m$ i.e., the replacement period. Noting the bottleneck of multiple mutation cycles in \model{}, we conducted an experiment to determine if a single mutation cycle could match the performance of Sample $N = 15$. In \Cref{tab:tuning}, we show the results of different tested values of $m$ ranging from 20 to 100, incrementing by 20 in each step. As we increase the replacement period, we observe a consistent improvement in the results up to $m = 80$. Beyond this point, the performance drops. This pattern could be the result of exploration-exploitation trade-off, where infrequent exploitation leads to a lower final score. Striking the right balance between both strategies is the key to yield optimal results. The performance gain from increasing $m$
 can be attributed to the increased stability of the reward calculation process associated with longer replacement periods. Although single-round mutation does not directly allow for iterative improvement, leveraging the RM score as guidance across multiple mutation cycles, it can still be beneficial if it outperforms other methods. 
\begin{table}[t]  
  \centering 
  \small
  \resizebox{\linewidth}{!}{
  \begin{tabular}{l|c|c}  
    \toprule  
    Setting &  LC & WR \\
    \midrule
   
    \model{} (\#MC=1, nb=1,$m$=20)  & 26.80     & 28.39    \\ 
    \model{} (\#MC=1, nb=1,$m$=40) & 27.01     & 28.33    \\  
    \model{} (\#MC=1, nb=1,$m$=60)  & 27.66     & 29.62    \\
    \model{} (\#MC=1, nb=1,$m$=80)  & 28.97     & 30.95      \\ 
    \model{} (\#MC=1, nb=1,$m$=100)  &  26.84   & 29.36        \\ 
    \bottomrule
  \end{tabular}  
  }
  \caption{The effect of changing replacement period, $m$, of \model{}.}  
  \label{tab:tuning}  
\end{table}

\paragraph{Ablations and Inference Time.}
We ablation study demonstrates that DARWIN needs a strong reward model to be effective. Additionally, we found that computing the reward score of each state with an extra look-ahead length does not improve performance. We also present the time complexity as well as the empirical inference time. The details to these are presented in Appendix \ref{ablations}.

\paragraph{Putting All of It Together.}
To summarize, we list all the key observations below:
\newtcolorbox{mychecklist}[1]{  
  colback=lightblue, 
  colframe=black, 
  sharp corners, 
  boxsep=5pt, 
  left=5pt, 
  right=5pt, 
  top=5pt, 
  bottom=5pt, 
  before=\vspace{10pt}, 
  after=\vspace{10pt}, 
  width=\linewidth, 
  #1 
} 
\begin{itemize}[label=\checkmark, wide, labelindent=0pt, itemsep=0pt, parsep=0pt, topsep=0pt]  
    \item \model{} outperforms strong baselines on Alpacaeval 2 and MT-Bench.
    \item \model{} when applied to LLMs aligned with preference modeling techniques improve their performance by 10-13\%.
    \item Tuning hyperparameters of \model{} is important. We found more mutation cycles and a larger replacement period are generally helpful.
    \item Look-ahead reward computation does not improve performance of \model{}.
    \item \model{} needs a strong reward model to excel.
  \end{itemize} 
  

\section{Related Works}

Inference-Time compute methods and LLM alignment are quite active research area aimed at aligning LLMs to human intentions, thus making them more useful. 
\paragraph{Alignment with Reinforcement Learning (RL).}
RL-based approaches~\cite{christiano2017deep} are shown to be effective in aligning LLMs with human preferences effectively. 
Models like Claude~\cite{bai2022training} and InstructGPT~\cite{ouyang2022training} use this technique, fitting a reward model to human preferences and optimizing the policy with Proximal Policy Optimization (PPO)~\cite{schulman2017proximal}. 

\paragraph{Alignment without Reinforcement Learning.}
The complexity and instability of RLHF have led to the development of alternative alignment methods such as 
DPO~\cite{rafailov2024direct}, ORPO~\cite{hong2024reference}, and SimPO~\cite{meng2024simpo}. Instead of relying on a trained reward model, these approaches rely on the LLM under training as the reward provider. 

\paragraph{Inference-time Alignment.}
Inference-time strategies like Augmented Recurrent Generation Strategies (ARGS)~\cite{args} and Rewindable Auto-regressive Inference Networks (RAIN)~\cite{li2024rain} offer innovative solutions. ARGS dynamically adjusts generation strategies to enhance output, while RAIN~\cite{li2024rain} employs a rewindable auto-regressive alignment technique to reduce harmful outputs at the token level without a reward model. Additionally, \citet{huang2024deal} propose DeAL, a heuristic-guided search process to improve adherence to alignment objectives during decoding.

\section{Conclusion}
In this work, we demonstrate the effectiveness using inference-time compute for general alignment. We proposed \model{}, a inference-time alignment technique that employs evolutionary strategies to implement exploration and exploitation aspects for a more-balanced reward optimization of the generated LLM responses. The empirical results strongly indicate our approach's supremacy over existing inference-time alignment methods and competitiveness with preference optimization methods. 

\section{Limitations}
The experiments conducted in this study utilized the Llama3-Instruct and Mistral-Instruct models with 8B and 7B parameters respectively. Due to computational limitations, the findings may not be applicable to models of larger sizes, as those experiments could not be performed. To enhance its inference speed, \model{} requires the implementation of inference time optimization techniques.

\section{Potential Risks}
Not applicable.

\section{Ethical Considerations}

Not applicable.

\bibliography{custom.bib}
\clearpage
\appendix
\begin{figure*}[ht]
    \centering
    \begin{subfigure}[b]{0.49\textwidth}
        \centering
        \includegraphics[width=\textwidth]
        {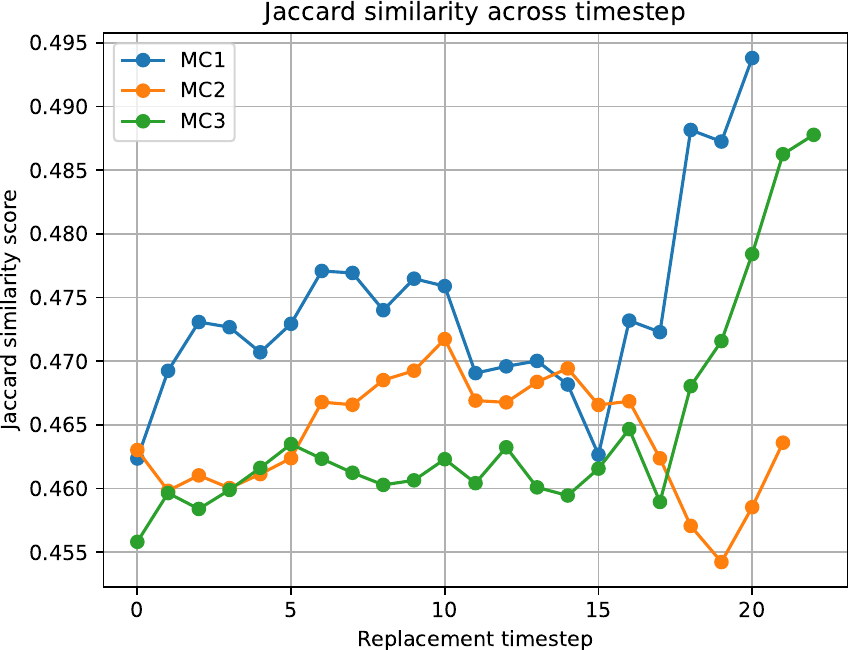}
        \caption{}
        \label{fig:jaccard_score}
    \end{subfigure}
    \hfill
    \begin{subfigure}[b]{0.49\textwidth}
        \centering
        \includegraphics[width=\textwidth]
        {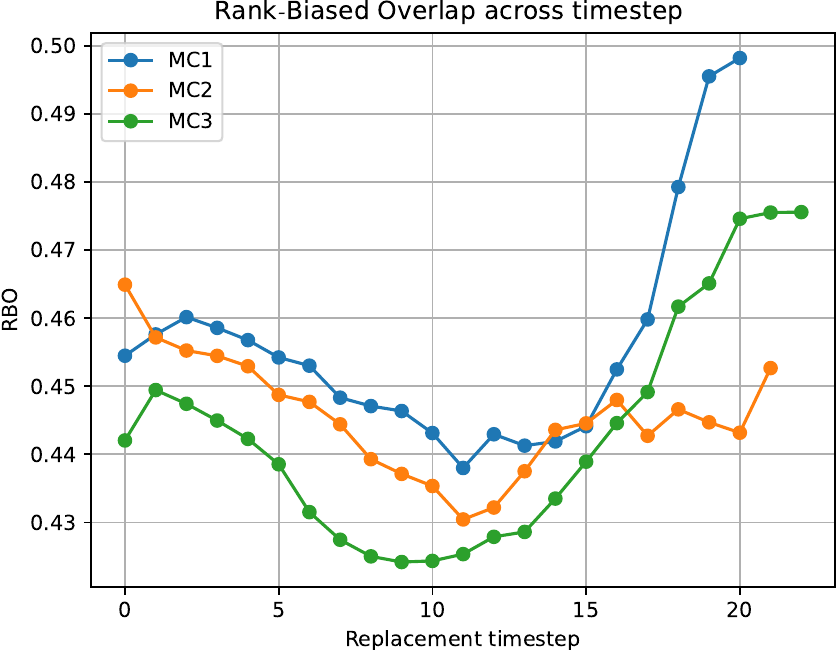}
        \caption{}
        \label{fig:rbo}
    \end{subfigure}
    \caption{An illustration of the average (a) \emph{Jaccard similarity} and (b) \emph{rank-biased overlap} (RBO) between the sets of the top-$k$ rewarded beams in two consecutive replacement cycles, where $k=3$.}
    \label{fig:top_k_sim}
\end{figure*}
\begin{figure*}[ht]
    \centering
    \begin{subfigure}[b]{0.3\textwidth}
        \centering
        \includegraphics[width=\textwidth]
        {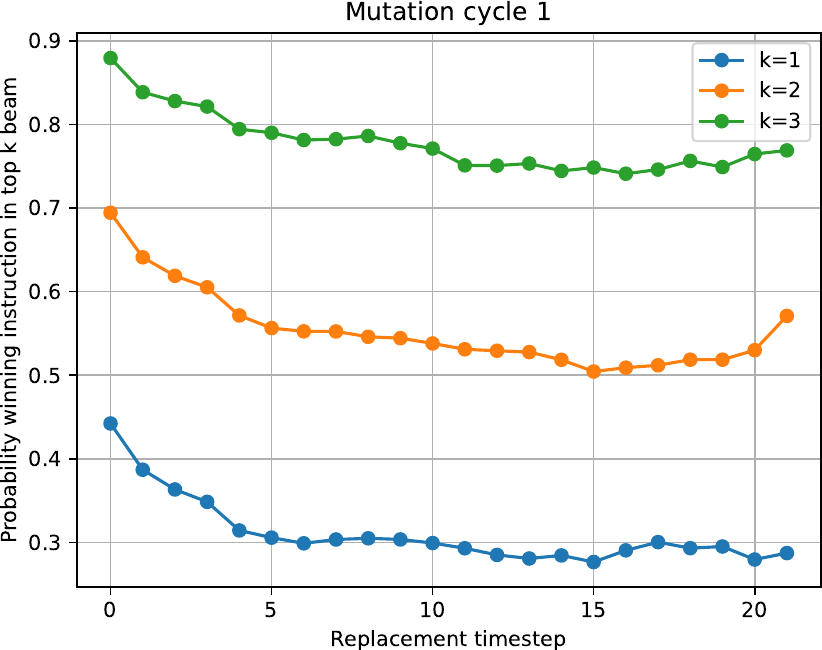}
        \caption{}
        \label{fig:top_k_mc1}
    \end{subfigure}
    \hfill
    \begin{subfigure}[b]{0.3\textwidth}
        \centering
        \includegraphics[width=\textwidth]
        {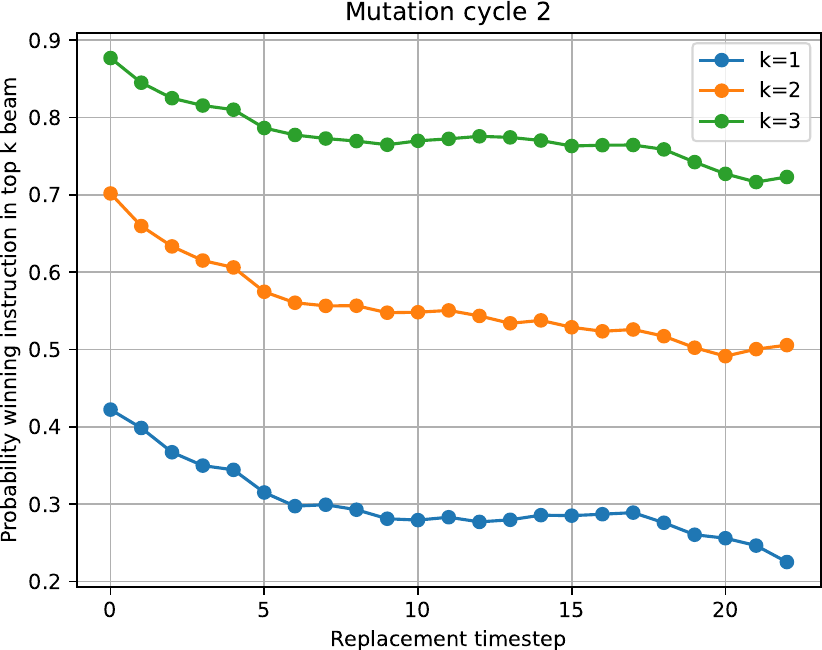}
        \caption{}
        \label{fig:top_k_mc2}
    \end{subfigure}
    \begin{subfigure}[b]{0.3\textwidth}
        \centering
        \includegraphics[width=\textwidth]
        {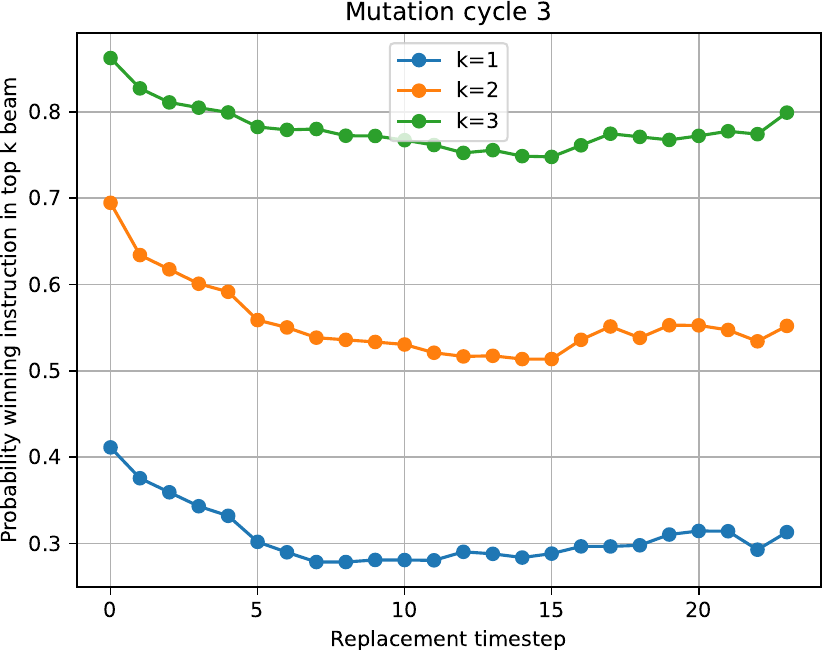}
        \caption{}
        \label{fig:top_k_mc3}
    \end{subfigure}
    \caption{A plot of the probability of the \emph{winning instruction} of a mutation cycle appearing with the top-k beams of a given replacement step in mutation cycle (a) 1, (b) 2, and (c) 3.}
    \label{fig:top_k_mc}
\end{figure*}

\section{Dynamics of Reward-guided Beam Replacement in \model{}}
\label{sec:beam-dynamics}

\noindent We want to investigate the behavior of reward-guided beam replacement of \model{}. We calculated the average \emph{Jaccard similarity} and \emph{Rank-Biased Overlap} (RBO) \cite{rbo} of top-k winning beams at each replacement step as shown in \cref{eq:jaccard_eq,eq:rbo_eq}, where the results are shown in \cref{fig:jaccard_score,fig:rbo} respectively. We also compute the average probability of final winning instruction $I_\text{win}$ in top-k rewarded beams at each replacement step for all mutation cycles shown in \cref{eq:ins}. The results are shown in \cref{fig:top_k_mc1} to \cref{fig:top_k_mc3}. All scores are computed with a smoothing average across 5 timesteps.
\\
\text{Jaccard Similarity between sets A and B:}
\begin{equation}
J(A, B) = \frac{|A \cap B|}{|A \cup B|}
\end{equation}
\\
\text{Average Jaccard Similarity at replacement step t:}
\begin{equation}
      \frac{1}{N} \sum_{i=1}^{N} J\left( I_{\text{top-}k}^{(t)}, I_{\text{top-}k}^{(t+1)} \right)
    \label{eq:jaccard_eq}
\end{equation}
\text{Rank-Biased Overlap between list $S$ and $T$ = RBO}
\begin{equation}
    (S,T,p) = (1-p) \sum_{d=1}^{\infty} p^{d-1} \cdot \frac{|S_{1:d} \cap T_{1:d}|}{d}
\end{equation}
\text{Average RBO at replacement step t}
\begin{equation}
      \frac{1}{N} \sum_{i=1}^{N} RBO\left( I_{\text{top-}k}^{(t)}, I_{\text{top-}k}^{(t+1)},p \right)
\label{eq:rbo_eq}
\end{equation}
\text{Average probability $I_{\text{win}}$ in $I_{\text{top-}k}^{(t)}$ }

\begin{equation}
     = \frac{1}{N} \sum_{t=1}^N \mathds{1}_{\{I_{\text{win}} \in I_{\text{top-}k}^{(t)}\}}
    \label{eq:ins}
\end{equation}

\paragraph{Reward-guided Beam Replacement Converges towards the Latter Replacement Steps.}
\cref{fig:top_k_sim} shows that both \emph{Jaccard similarity} and \emph{rank-based overlap} have a noticeably-increasing trend towards the latter replacement timesteps across all the mutation cycles. This suggests that replacements become more stable in the latter replacement cycles, with relatively less variation between consecutive top-$k$ winning beams, indicating convergence. 

\paragraph{Collaboration through Reward-guided Beam Replacement.}
From our main result, the full \model{} outperforms \model{} without replacement across all mutation cycles on both WR and LC, indicating the effectiveness of replacements. \cref{fig:rbo} suggests that most of the collaboration across different beams occurs in the middle replacement cycles (roughly from 5 to 15 in \cref{fig:top_k_sim})---the decrease in RBO values in the middle cycles indicates rapid change in the ranked order of top-k rewarded beams across cycles, displaying no consistent pattern.  This variability suggests that the highest level of inter-beam collaboration occurs in this stage. As the algorithm converges, the RBO value starts to increase, indicating less variability in the top-k rewarded beams and, hence, less collaboration.

\paragraph{First Few Replacement Cycles are the Most Important.}
The \emph{winning instruction} of a mutation cycle is defined as the instruction that emerges the most among the top-k rewarded beams across the replacement cycles of a mutation cycle. We investigate the average probability of these instructions appearing in the top-k rewarded beams of each replacement step for $k=1,2,3$. We plot this average probability across three mutation cycles in \cref{fig:top_k_mc1,fig:top_k_mc2,fig:top_k_mc3}; we apply a smoothing average across five steps. The initial replacement step appears to be the most critical in determining the final winning instruction. This is evidenced by the highest probability of the winning instruction being among the top-k rewarded beams at the initial timestep. The probability decreases as timestep increases, signifying that the latter replacement steps have lesser significance in determining the final winning instruction. This also demonstrates that the importance of individual instruction decreases with each replacement timestep and collaboration becomes more important. One possible reason behind this phenomenon is as the sequence grows in length, the influence of small variations in the initial instructions diminishes. The previously generated context through replacement becomes increasingly dominant in guiding the algorithm's subsequent outputs, potentially overshadowing slight differences in each beam's guiding instruction.

\subsection{Overall Behaviour of \model{}}
\model{}'s behavior within a mutation cycle seems to have three phases: (i) \emph{Early Phase} -- Initial replacement steps are important. They begin steering states towards the eventual winning instruction, setting the overall direction; 
(ii) \emph{Middle Phase} -- Characterized by rapid collaboration across beams. This phase focuses on refining and improving the generated sequences through information exchange;
(iii) \emph{Late Phase} -- Generation continues with relatively less collaboration across beams, as \model{}'s behavior starts to resemble Best-of-N approach.

\section{Analyses} \label{ablations}
\paragraph{Inference Time and Setting.}
In \Cref{tab:inference-time}, we compare the inference time of Sample-$N$ \& Best-of-$N$ with \model{}. We use one A100 with 80GB of GPU memory for this experiment. We note that Best-of-N 5, and Best-of-N 15 has similar inference time due to parallelizing inference at the cost of more VRAM.
All inferences are done at half precision. The generation hyperparameters are set to temperature = $0.7$, max new tokens = $2048$, and top \(k\) = $40$. We did not perform any tuning of the generation hyperparameters. Additionally, we use $n = 5$ for the number of mutations and the replacement period of $m = 40$ tokens in all our experiments. The value of $k$ in choosing top-$k$ mutations was set to 3.

\paragraph{Asymptotic Time Complexity.}
The time complexity of our methods are presented with the following assumptions: (i) the generation LLM has a time complexity of  $\mathcal{O}(L^2)$, where $L$ is the generated sequence length, assuming the prior token key-value matrices are properly cached. (ii) the reward model has a time complexity of $\mathcal{O}(k)$ where \(k\) is the input sequence length, and (iii) the number of transformer layers in an LLM is constant. Assuming $m$ is the replacement period, there will be $L/m$ replacement cycles. Let $n$ denote the number of beams in the generation. The time complexity of reward calculation across all replacement is $\sum_{s=1}^{s=\frac{L}{m}} \mathcal{O}(nsm) = \mathcal{O}(nm(\frac{L}{m})(1+\frac{L}{m})/2) = \mathcal{O}(\frac{nL^2}{m})$. Therefore the time complexity of reward-guided beam search generating length $L$ and replacement period of $m$ is given by $\mathcal{O}(\frac{nL^2}{m}+nL^2) = \mathcal{O}(nL^2)$. Sample $N$ \& Best-of-$N$ has a complexity of $\mathcal{O}(nL^2)$ as well.
For \model{}, each time we perform a replacement, the previous key-value caching assumption becomes invalid due to different instructions for each beam. Consequently, we have to recompute the past key-value every replacement. The complexity for this operation is $\sum_{s=1}^{s=\frac{L}{m}} \mathcal{O}(ns^2m^2) = \mathcal{O}(nm^2(\frac{L^3}{m^3})) = \mathcal{O}(\frac{nL^3}{m})$. Hence, the final time complexity for \model{} is $\mathcal{O}(\frac{nL^3}{m})$. 

\paragraph{Weak vs Strong Reward Model (RM).}

In this study, we demonstrate that the choice of reward model significantly impacts the WR and LC performance (\Cref{tab:RM}) when used as heuristics to guide the exploration process. We compare two reward models: the Llama3-based 8B RM as explained above and a smaller RM based on Gemma, a 2B parameter-sized model. Our findings indicate that a weaker RM may provide noisy heuristics for the exploitation techniques, leading to poor results. Notably, when using the smaller RM, \model{} performs worse than the Sample $N$ \& Best-of-$N$ baseline. These results suggest that \model{} requires a strong RM to outperform the baselines effectively as a weak RM might provide noisy reward scores for truncated responses during replacement-based exploitation. With a strong RM, both these approaches improve their performance the improvement is more prominent with \model{} as it gains about 6\% for both LC and WR.

\begin{table}[t]  
  \centering 
  \begin{minipage}{.5\textwidth}
  \tiny
  \resizebox{\linewidth}{!}{
  \begin{tabular}{l|c}
    \toprule  
    Setting &  Inference Time (sec.) \\
    \midrule
    Sample-$N$ \& Best-of-$N=5$ & 14.6 \\
    Sample-$N$ \& Best-of-$N=15$ & 18.5 \\
    \midrule
    \model{} (\#MC=1, $m$=40, $l=0$)  & 33.9        \\ 
    \model{} (\#MC=2, $m$=40, $l=0$)  &  62.2         \\ 
    \model{} (\#MC=3, $m$=40, $l=0$)  &  85.1        \\ 
    \midrule
    \model{} (\#MC=1, $m$=40, $l=25$)  & 46.1        \\ 
    \model{} (\#MC=2, $m$=40, $l=25$)  & 81.0         \\ 
    \model{} (\#MC=3, $m$=40, $l=25$)  & 111.2      \\ 
    \bottomrule
  \end{tabular}  
  }
  \end{minipage}
  \caption{Inference time comparison. $l$ denotes the look-ahead length for reward computation.}  
  \label{tab:inference-time}  
\end{table}

\begin{table}[ht!]
  \centering
  \scriptsize 
  \begin{tabular}{l|cccccc}
    \toprule
    \multirow{2}{*}{Setting} & \multirow{2}{*}{RM} & \multicolumn{2}{c}{Llama3} & \multicolumn{2}{c}{Mistral2} \\ \cline{3-6}
                             &                     & LC & WR & LC & WR \\ \hline
    Sample-5 \& Best-of-5 & Gemma-2B  & 25.62 & 25.54 & 20.94 & 19.15 \\
    Sample-5 \& Best-of-5 & Llama3-8B  & 26.63 & 26.66 & 23.87 & 20.86 \\
    \midrule
    \model{} (\#MC=1) & Gemma-2B  & 21.15 & 22.69 & 17.86 & 17.00 \\
    \model{} (\#MC=1) & Llama3-8B  & 27.02 & 28.33 & 24.42 & 21.85 \\
    \bottomrule
  \end{tabular}
  \caption{Impact of weak (Gemma-2B) and strong (Llama3-8B) RMs tested with Llama3-Instruct and Mistral-Instruct.}
  \label{tab:RM}
\end{table}

\paragraph{Why to Choose top-$k$ Mutated Instructions?}

We want to investigate the necessity of selecting multiple top-$k$ mutated instructions, rather than consistently opting for the single most impactful one. Our decision to set $k$ greater than 1 was influenced by principles from reinforcement learning-based search strategies. These strategies suggest that allowing for some exploration, by considering actions with lower immediate rewards, can potentially lead to higher overall rewards in the long run. This approach acknowledges that some actions, while seemingly less beneficial at time $t$, may ultimately prove more valuable at time $t + n$. We present our findings of this experiment in \Cref{tab:top-k}. The results indicate that setting $k$ to a value greater than 1 enhances overall performance, thus validating our approach of considering multiple top candidates rather than focusing solely on the single best option. \cref{sec:beam-dynamics} further delves into the dynamics of these top-$k$ beams.
\begin{table}[t]  
  \centering 
  
   \begin{minipage}{.4\textwidth}
   \tiny
  \resizebox{\linewidth}{!}{
  \begin{tabular}{l|c|c}  
    \toprule  
    Setting &  LC & WR \\
    \midrule
    \model{} (\#MC=1, $k$=1)  & 26.17     & 28.16    \\ 
    \model{} (\#MC=2, $k$=1) & 27.69     & 30.85    \\  
    \model{} (\#MC=3, $k$=1)  & 28.54     & 32.65    \\
    \midrule
    \model{} (\#MC=1, $k$=3)  & 27.01     & 28.33      \\ 
    \model{} (\#MC=2, $k$=3)  &  28.70   & 31.47        \\ 
    \model{} (\#MC=3, $k$=3)  &  30.47   & 33.90        \\ 
    \bottomrule
  \end{tabular}  
  }
  \end{minipage}

  \caption{Understanding whether choosing top-$k$ mutated instructions is needed.}  
  \label{tab:top-k}  
\end{table}

\begin{figure*}[h]
    \centering
    \includegraphics[width=\textwidth]{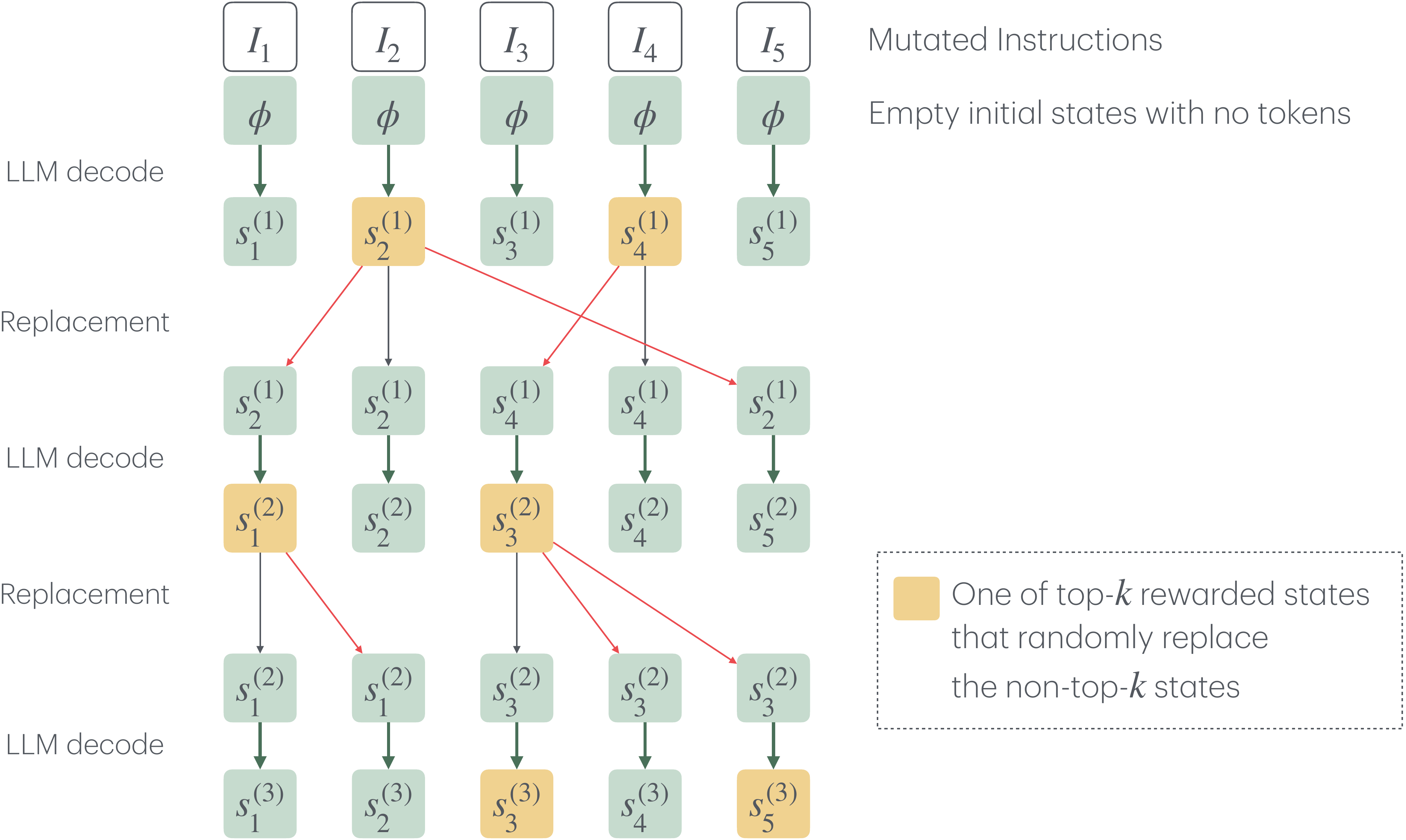}
    \caption{The tree search conducted by \model{} to obtain an aligned output.}
    \label{fig:darwin-tree}
\end{figure*}

\paragraph{Look-ahead Reward does not Improve \model{}.}
In \model{}, we focus on the reward score of the current states $s_i^{tm}$, where $t$ represents the number of replacement cycles and $m$ denotes the replacement period. However, as noted by \citet{huang2024deal}, calculating state rewards based on the future tokens could be beneficial, as reward models are trained on complete responses rather than partial or truncated ones.
We are therefore interested in evaluating our model's performance when the reward score $\mathcal{R_\theta}$ is calculated using $s_i^{(tm+l)}$, where $l$ represents the look-ahead length. This approach involves generating an additional $l$ tokens after producing $tm$ tokens in a replacement cycle $t$. We then use this extended sequence to compute the reward, $\mathcal{R_\theta}(s_i^{(tm+l)}, I)$, which guides the selection of the most promising beams of length $tm$ to replace others. 
Note that the look-ahead reward computation introduces a computational overhead, slowing down \model{}. Contrary to the findings of \cite{huang2024deal}, our experimental results, presented in \cref{tab:look-ahead}, show that incorporating look-ahead rewards does not improve \model{}'s effectiveness. We tried varied look-ahead sizes for these experiments such as $l = 25, 50, \text{and } 100$. The look-ahead length of $l=25$ slightly harms the performance. This indicates that the look-ahead rewards do not have much importance.
\paragraph{Parallels with Tree Search.}
\cref{fig:darwin-tree} presents the search process of \model{} as a fixed-width tree search, where the number of mutations defines the treewidth. The pruning operation is represented by top-k state/node selection followed by a state-replacement operation. This pruning operation eliminates potentially unfruitful states while combining the influence of two distinct instructions in the subsequent decoding. Thus, the pruning operation allows both exploitation and exploration of states. Another notable difference with a general tree search is the lack of branching from a state. The only branching is achieved via the replacement operation which essentially duplicates the top states. In contrast, \citet{huang2024deal} explores multiple branches at the token level to optimize reward, although it lacks any recombination across multiple instructions.

\begin{table}[t]  
  \centering 
  	
  \begin{minipage}{.5\textwidth}
   \tiny
  \resizebox{\linewidth}{!}{
  \begin{tabular}{l|c|c}  
    \toprule  
    Setting &  LC & WR \\
    \midrule
    \model{} (\#MC=1, $m$=40, $l$=0)  & 27.01     & 28.33      \\ 
    \model{} (\#MC=2, $m$=40, $l$=0)  &  28.70   & 31.47        \\ 
    \model{} (\#MC=3, $m$=40, $l$=0)  &  30.47   & 33.90        \\ 
    \midrule
    \model{} (\#MC=1, $m$=40, $l$=25)  &  25.97   & 28.11    \\ 
    \model{} (\#MC=2, $m$=40, $l$=25)  & 27.66   & 31.04    \\ 
    \model{} (\#MC=3, $m$=40, $l$=25)  & 28.62  & 33.10    \\
    \arrayrulecolor{black!50}\specialrule{1pt}{1\jot}{1\jot}
    \model{} (\#MC=1, $m$=80)  & 28.97     & 30.95      \\ 
    \model{} (\#MC=1, $m$=80, $l$=25)  & 25.73     & 28.07    \\ 
    \model{} (\#MC=2, $m$=80, $l$=25)  & 27.88     & 31.62    \\ 
    \model{} (\#MC=3, $m$=80, $l$=25)  & 28.49     & 33.56    \\
    \midrule
    \model{} (\#MC=1, $m$=40, $l$=50)  & 26.68     & 29.53    \\ 
    \model{} (\#MC=1, $m$=40, $l$=100)  & 26.10     & 29.15    \\
    \bottomrule
  \end{tabular}  
  }
  \end{minipage}
  \caption{The impact of look-ahead reward calculation. The non-zero value of $l$ indicates the experiment was conducted with the look-ahead reward computation.}  
  \label{tab:look-ahead}  
\end{table}

\subsection{Detailed Illustration of the Steps of \model{}}
\begin{figure*}[ht]
    \centering
    \includegraphics[width=0.9\textwidth]
    {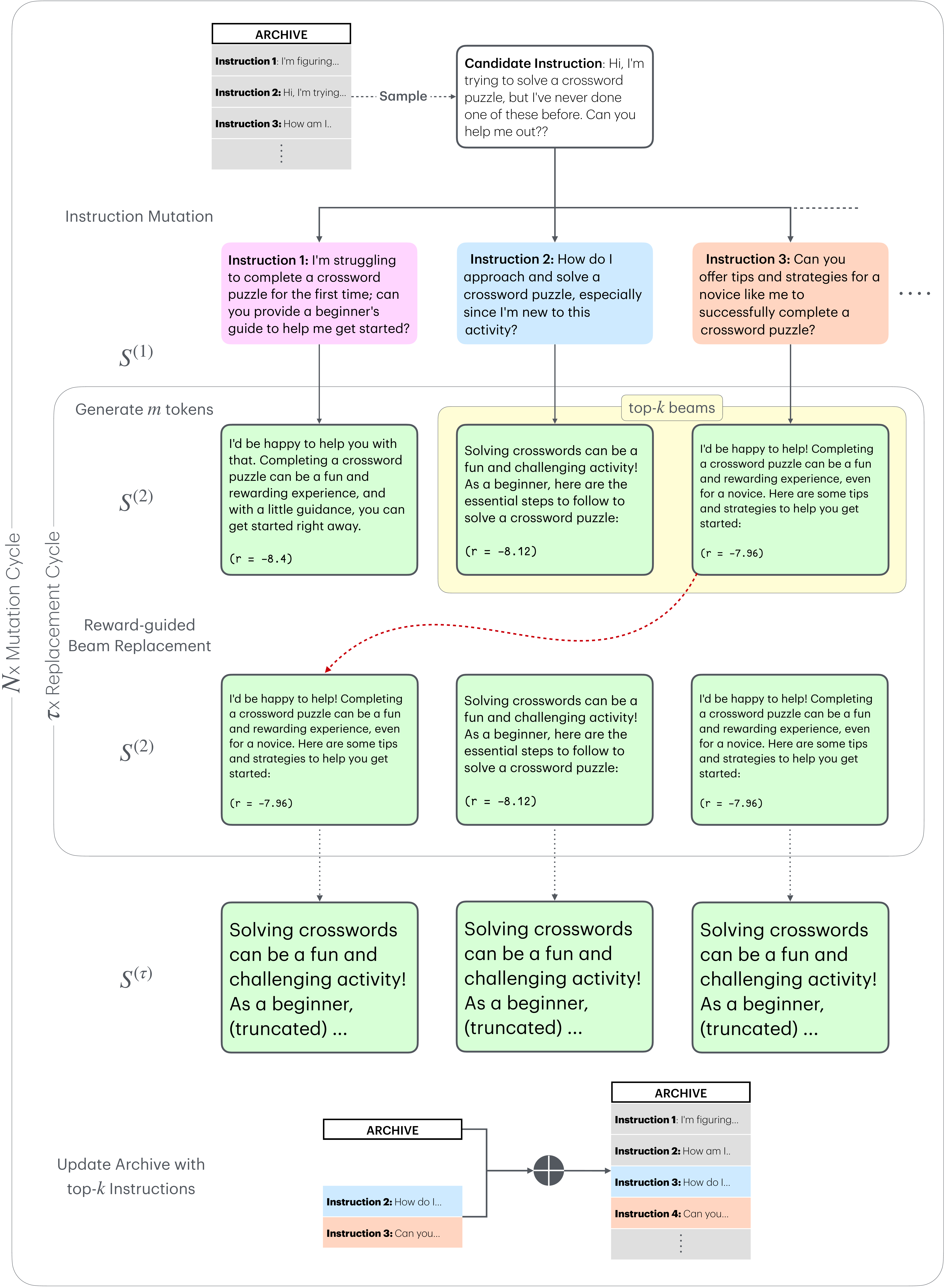}
    \caption{An illustration of the steps in a mutation cycle of \model{}. At each mutation cycle, a candidate instruction is sampled from the archive and mutated into n instructions. Reward-guided replacement is performed for every $m$ tokens until all the states have reached the end. The top-$k$ instruction is computed and updated in the archive, replacing the candidate instruction if the new output receives a higher reward.}
    \label{fig:darwin_detail}
\end{figure*}

A demonstration of the steps of \model{} is shown in \Cref{fig:darwin_detail}.

\onecolumn

\section{Sample Output of \model{}}
A sample output of \model{} using Llama3-Instruct.

\begin{tabularx}{\textwidth}{|>{\hsize=.3\hsize}X|>{\hsize=1.7\hsize}X|}
\toprule  
\textbf{Stage} & \textbf{Textual Output} \\  
\midrule  
\rowcolor[gray]{.9}  
\textsc{Prompt} & Write a social media post about the call for collaboration on a crowdsourcing project in a persuasive way. \\ 
\midrule  
\rowcolor{lightyellow}  
\textsc{Mutation Cycle 1} &
EXCITING OPPORTUNITY ALERT! We're on a mission to create something BIG and we need YOUR help! We're launching a crowdsourcing project to [briefly describe the project and its purpose], and we're calling on all creative minds to join forces and make it happen! As a valued member of our community, we're inviting you to be a part of this groundbreaking initiative. Whether you're a designer, writer, developer, or simply a passionate individual with a great idea, we want to hear from you! By collaborating with us, you'll have the chance to: Contribute to a project that will make a real impact Connect with like-minded individuals from around the world Develop your skills, and showcase your talents Be part of a dynamic and supportive community But that's not all! By participating in this project, you'll also get: Recognition for your work and contributions Access to exclusive updates and behind-the-scenes content A chance to be part of a project that will be showcased to a global audience So, what are you waiting for? Join our crowdsourcing project today and be a part of something truly special! Click the link below to learn more and get started: [Insert link to project page or sign-up form] Don't miss out on this incredible opportunity to make your mark and be part of something BIG! Share with your friends and networks to spread the word! \#CrowdsourcingProject \#Collaboration \#Innovation \#Creativity \#Impact \#CommunityFirst \#MakeYourMar \\ 
\midrule  
\rowcolor[gray]{.9}  
\textsc{Mutation Cycle 2} & **Join the Movement: Amplify Your Impact through Collaboration!** Are you passionate about making a difference in your community or industry? Do you have a great idea that could change the world? We're excited to announce a unique opportunity to collaborate with like-minded individuals and organizations to create something truly remarkable! **What is this all about?** We're launching a crowdsourcing project that brings together experts, innovators, and changemakers to tackle some of the world's most pressing challenges. Our goal is to harness the collective power of our community to drive positive change and create innovative solutions that benefit everyone. **How can you get involved?** Join our project as a contributor, and you'll have the chance to: Collaborate with a diverse group of talented individuals from around the globe Share your skills, expertise, and ideas to co-create something amazing Be part of a dynamic community that's shaping the future Get recognized for your contributions and showcase your work to a global audience **What kind of projects can you participate in?** We're open to any idea that has the potential to make a positive impact. Some examples include: Sustainable energy solutions Mental health initiatives Environmental conservation efforts Education and skills development programs Innovative technologies for social good **How does it work?** 1. Submit your project idea or join an existing one that resonates with you. 2. Collaborate with our community to refine your idea and create a plan. 3. Contribute your skills and expertise to bring your project to life. 4. Share your progress and results with the world through our social media channels. **Ready to join the movement?** Click the link below to learn more and get started! [Insert link to project page or sign-up form] **Let's make a difference, together!** Share this post with your network and help spread the word about our crowdsourcing project. Together, we can achieve amazing things! \#Crowdsourcing \#Collaboration \#Innovation \#SocialImpact \#MakeADifference \\ 
\midrule  
\rowcolor{lightyellow}  
\textsc{Mutation Cycle 3} & **Unlock Your Creativity and Make a Difference!** Are you ready to unleash your inner innovator and contribute to a project that can change the world? We're excited to announce the launch of our latest initiative, "CrowdSpark," a platform that empowers individuals to come together and co-create innovative solutions to real-world problems. By joining CrowdSpark, you'll be part of a community that values collaboration, creativity, and inclusivity. Our platform is designed to bring people together from diverse backgrounds and expertise, allowing you to learn from each other, share your ideas, and build something amazing. Here are just a few reasons why you should join us: **Amplify Your Impact**: By working together, we can achieve more than we could alone. Our collective efforts can lead to groundbreaking discoveries, new perspectives, and meaningful solutions that make a real difference. **Discover New Connections**: Meet like-minded individuals who share your passions and interests. Build lasting relationships, learn from each other's experiences, and grow as a community. **Unleash Your Creativity**: Our platform is designed to inspire and empower you to think outside the box. Share your ideas, get feedback, and watch your concepts come to life. **Be Part of Something Bigger Than Yourself**: Join a movement that's all about creating positive change. Together, we can tackle complex challenges, drive innovation, and make the world a better place. Ready to get started? Here's how you can participate: 1. Follow us on social media to stay updated on the latest news and project updates. 2. Register for our platform to create your profile, share your ideas, and connect with other members. 3. Join our community forums to discuss topics, share resources, and get feedback on your projects. Don't miss this opportunity to be part of something amazing! Join CrowdSpark today and let's make a difference together! **\#CrowdSpark \#Teamwork \#Innovation \#Collaboration \#Creativity \#PositiveChange** \\ 
\bottomrule  
\end{tabularx} 
\section{Mutator-Prompt Template}
\label{sec:mutator-prompt}

\begin{tabular}{>{\raggedright\arraybackslash\tt}p{0.94\linewidth}<{}}
You are a professional prompt engineer. You are given an original instruction and your goal is to mutate the instruction into 5 different instruction that will improve the clarity of original instruction. The mutated instruction should not deviate from the original instruction and they should provide the same general intention.

Hint: Think of adding more details,removing details in the instruction or change certain phrasing when mutating the instruction.
Only give the mutated instruction in a list order.
Original instruction: How to make a cake?
1. How to bake a delicious cake?
2. Step-by-step guide to making a perfect cake from scratch
3. How to bake a cake?
4. Detailed instructions for creating a professional-quality cake at home
5. How to prepare a beautiful homemade cake?
Original instruction: |\exprompt{}|
\end{tabular}

\end{document}